\documentclass[lettersize,journal]{IEEEtran}

\usepackage{amsmath,amsfonts}
\usepackage{booktabs}
\usepackage[table]{xcolor}
\usepackage{threeparttable}
\usepackage{amssymb}   
\usepackage{pifont}    

\newcommand{\cmark}{\checkmark}
\newcommand{\xmark}{\ding{55}} 
\usepackage{makecell}
\usepackage{multirow}
\usepackage{multicol}
\usepackage{xspace}
\usepackage{bm}
\usepackage{algorithmic}
\usepackage{algorithm}
\usepackage{array}
\usepackage[caption=false,font=scriptsize,labelfont=sf,textfont=sf]{subfig}
\usepackage{textcomp}
\usepackage{stfloats}
\usepackage{url}
\usepackage{verbatim}
\usepackage{graphicx}
\usepackage{cite}
\usepackage{bbding} 
\usepackage{pifont} 
\usepackage{arydshln} 
\usepackage{graphicx} 
\usepackage[colorlinks,linkcolor=blue,citecolor=black]{hyperref}
\usepackage{cite}
\usepackage{ifthen}

\newboolean{anonymizeauthors}
\setboolean{anonymizeauthors}{false}
\makeatletter
\let\NAT@parse\undefined
\makeatother
\usepackage[colorlinks,linkcolor=blue,citecolor=black]{hyperref}
\makeatletter
\DeclareRobustCommand\onedot{\futurelet\@let@token\@onedot}
\def\@onedot{\ifx\@let@token.\else.\null\fi\xspace}
 
\def\ie{\emph{i.e}\onedot}

\makeatother

\newcommand{\degcm}[2]{{#1}$^\circ${#2}cm}
\newcommand{\sepdegcm}[2]{$({#1}^\circ,\,{#2}\,\mathrm{cm})$}

\usepackage[capitalize]{cleveref}
\crefname{section}{Sec.}{Secs.}
\Crefname{section}{Section}{Sections}
\Crefname{table}{Table}{Tables}
\crefname{table}{Tab.}{Tabs.}
\graphicspath{{Fig/}}
\begin{document}
	
	\title{Native-Domain Cross-Attention for Camera–LiDAR Extrinsic Calibration Under Large Initial Perturbations}
	\ifthenelse{\boolean{anonymizeauthors}}{
		\author{Anonymous Authors%
			\thanks{This version has been anonymized for review.}
		}
	}{
		\author{Ni Ou$^1$, Zhuo Chen$^2$, Xinru Zhang$^3$ and Junzheng Wang$^{1,*}$
			\thanks{$^{1}$Ni Ou, Junzheng Wang are with the School of Automation, Beijing Institute of Technology, Beijing, 100081, China.}
			\thanks{$^{2}$Zhuo Chen is with the Robot Perception Lab, Centre for Robotics Research, Department of Engineering, King's College London, London WC2R 2LS, United Kingdom.}
			\thanks{$^{3}$Xinru Zhang is with the School of Integrated Circuits and Electronics, Beijing Institute of Technology, Beijing, 100081, China.}
			\thanks{$^*$This work was supported by the National Natural Science Foundation of China under Grant 62173038. Corresponding Author: Junzheng Wang. Email: {\tt\small wangjz@bit.edu.cn}}
		}
	}
	
	\markboth{This paper has been accepted to IEEE Robotics and Automation Letters.}%
	{Ni Ou \MakeLowercase{\textit{et al.}}: Native-Domain Cross-Attention for Camera–LiDAR Extrinsic Calibration}
	
	\IEEEpubid{0000--0000/00\$00.00~\copyright~2021 IEEE}
	
	\maketitle
	
	\begin{abstract}
		Accurate camera–LiDAR fusion relies on precise extrinsic calibration, which fundamentally depends on establishing reliable cross-modal correspondences under potentially large misalignments. Existing learning-based methods typically project LiDAR points into depth maps for feature fusion, which distorts 3D geometry and degrades performance when the extrinsic initialization is far from the ground truth. To address this issue, we propose an extrinsic-aware cross-attention framework that directly aligns image patches and LiDAR point groups in their native domains. The proposed attention mechanism explicitly injects extrinsic parameter hypotheses into the correspondence modeling process, enabling geometry-consistent cross-modal interaction without relying on projected 2D depth maps. Extensive experiments on the KITTI and nuScenes benchmarks demonstrate that our method consistently outperforms state-of-the-art approaches in both accuracy and robustness. Under large extrinsic perturbations, our approach achieves accurate calibration in 88\% of KITTI cases and 99\% of nuScenes cases, substantially surpassing the second-best baseline. We have open sourced our code on \href{https://github.com/gitouni/ProjFusion}{GitHub} to benefit the community.
	\end{abstract}

	
	\section{Introduction}
	\label{sec.introduction}
	\IEEEPARstart{C}{amera} and LiDAR are indispensable sensors in autonomous driving, each providing unique yet complementary information essential for robust perception. Cameras deliver high-resolution imagery rich in semantic detail, whereas LiDAR offers precise structural information through three-dimensional, albeit sparse, point cloud measurements. The fusion of these modalities has substantially advanced intelligent transportation systems, enabling superior performance across a wide range of autonomous driving tasks, including object detection~\cite{Object-Detection1, Object-Detection2} and tracking~\cite{Object-Tracking1,Object-Tracking2}, SLAM~\cite{SLAM1, SLAM2}, and scene flow estimation~\cite{CamLiFlow,Flow2}.
	
	\begin{figure}[t]
		\centering
		\subfloat[Miscalibrated projection and depth map.]{
			\includegraphics[width=0.95\linewidth]{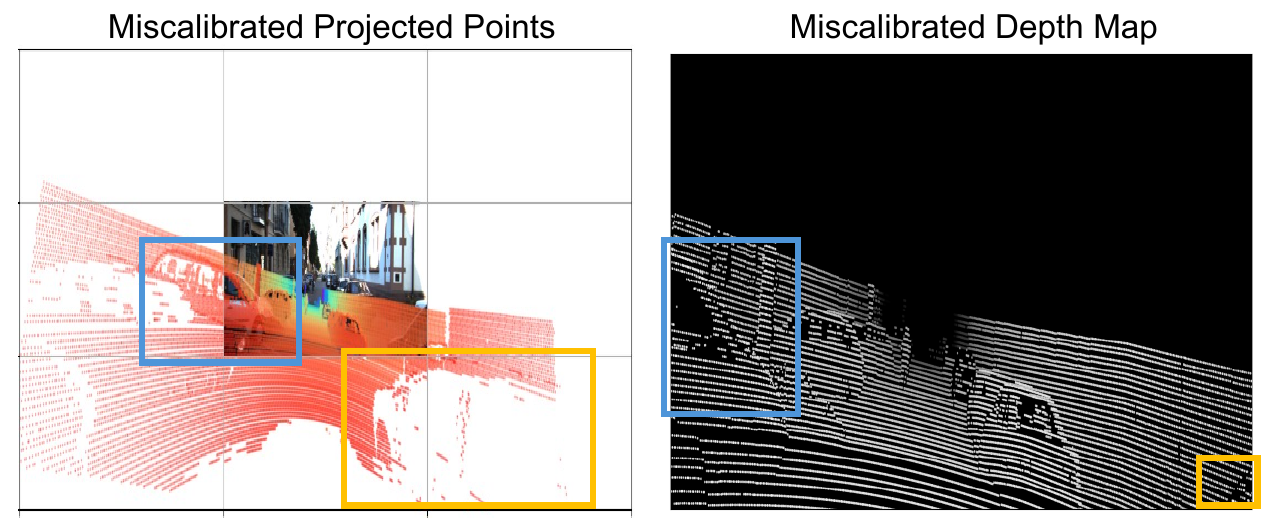}
			\label{Fig.abstract_depth_proj}
		}\\[4pt]
		\subfloat[Extrinsic-aware cross-attention.]{
			\includegraphics[width=0.95\linewidth]{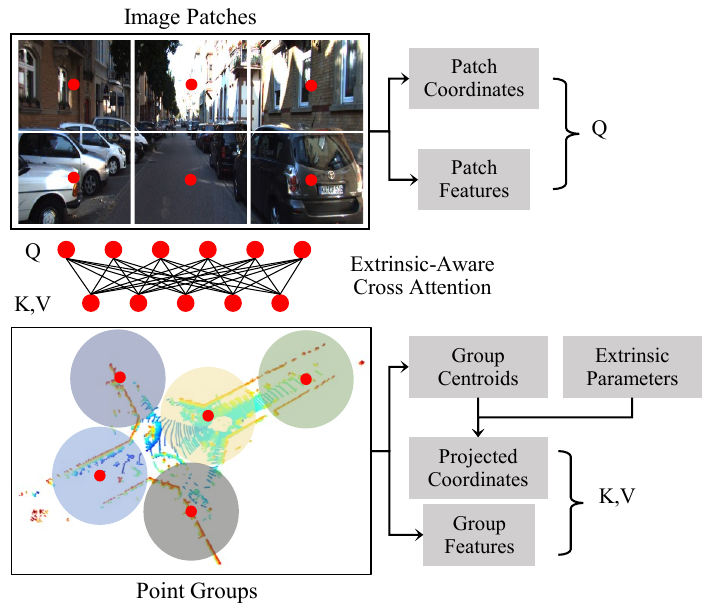}
			\label{Fig.abstract_pipeline}
		}
		\caption{
			Illustration of the problem caused by miscalibrated depth maps and our proposed method. 
			(a) Projecting LiDAR points onto the image plane produces incomplete and distorted object structures. 
			(b) Image patches and point groups are encoded separately and fused through extrinsic-aware cross-attention, enabling structure-preserving cross-modal feature interaction.
		}
		\label{Fig.abstract}
	\end{figure}

	Accurate fusion of camera and LiDAR data requires an extrinsic matrix that defines the spatial relationship between the two sensors. This matrix is typically obtained through a camera–LiDAR calibration process. The core challenge in calibration lies in extracting and matching reliable correspondences between camera and LiDAR observations. To facilitate this process, target-based methods have been developed using calibration objects such as planar boards~\cite{Calib-board,Calib-checkboard-passthrough} and boxes~\cite{Calib-Box}, which contain hand-crafted geometric features recognizable by both sensors. However, these methods require placing the calibration target at multiple positions in front of the sensors, making them impractical for online or in-vehicle calibration scenarios where extrinsic parameters may drift due to vehicle vibrations, thermal expansion, or gradual deformation of mechanical structures.
	
	\IEEEpubidadjcol
	To overcome these limitations, targetless calibration methods have been proposed to eliminate the reliance on dedicated calibration targets and instead exploit cross-modality correspondences present in natural scenes. Some approaches estimate the extrinsic matrix by applying the Perspective-n-Point (PnP) algorithm to matched correspondences, such as edges~\cite{Calib-edge1,Calib-pixellevel} or learned feature pairs~\cite{CorrI2P,CoFiI2P}. These methods offer strong interpretability and can perform well even under large initial calibration errors. However, their effectiveness is highly dependent on the recall of feature matching, rendering them sensitive to scene structure and environmental variations. In contrast, other works adopt an end-to-end learning framework~\cite{CalibNet,RGGNet,LCCNet,LCCRAFT,MSANet,Calibformer} that directly regresses extrinsic parameters from fused RGB images and miscalibrated LiDAR depth maps. Although these approaches avoid explicit correspondence extraction and matching, they often exhibit degraded performance when confronted with large initial calibration errors.
	
	We attribute the limitations of end-to-end calibration methods to their reliance on miscalibrated depth maps and fusion mechanism. As illustrated in \cref{Fig.abstract_depth_proj}, generating such a depth map requires projecting 3D LiDAR points onto a 2D grid using the given initial extrinsic matrix. Since projection is not a distance-preserving transformation, it distorts object geometry and inevitably discards points that fall outside the image frame, which hinders reliable feature extraction from LiDAR projections and weakens the effectiveness of subsequent fusion modules. To mitigate these issues, we redesign the point feature extraction branch and introduce a cross-modality fusion module. As shown in \cref{Fig.abstract_pipeline}, position-aware image and point features are extracted in their native domains, enabling cross-modality interaction while preserving floating-point precision and avoiding the discard of LiDAR points. Our main contributions are summarized as follows:
	
	\begin{itemize}
		\item We propose a novel end-to-end camera–LiDAR calibration framework that incorporates extrinsic-aware cross-attention, mitigating geometric distortion and point dropout introduced by depth-map projection, thereby enabling more reliable fusion of image and point features.
		
		\item We introduce a cross-modal coordinate alignment strategy that fundamentally differentiates our cross-attention from existing baselines. By injecting aligned coordinates alongside a harmonic embedding scheme, our mechanism expands the effective field of view and maintains high positional sensitivity, enabling robust feature correlation even under large initial perturbations.
		
		\item Extensive experiments on the KITTI~\cite{KITTI} and nuScenes~\cite{nuScenes} datasets demonstrate that our method consistently outperforms state-of-the-art baselines in both accuracy and robustness. Comprehensive ablation studies validate the effectiveness of each component of our approach.
	\end{itemize}
	
	\section{Related Works}
	\subsection{Target-based Methods}
	\label{subsec.target-based-methods}

	Target-based methods rely on calibration targets jointly observable by cameras and LiDARs to provide reliable geometric constraints. Planar chessboards~\cite{Calib-unni-chessboard, Calib-chessboard-plane-only,Calib-chessboard-line-plane} are widely used due to their point, line, and plane constraints. Variants such as triangular boards~\cite{Calib-triangular, Calib-triangular2} and circular-hole designs~\cite{L2v2t2, Calib-Cycle-Hole} further improve LiDAR feature extraction, especially for low-resolution sensors.

	Beyond planar designs, 3D calibration targets exploit richer geometric structures and are more readily available in natural environments. For example, V-shaped objects~\cite{Calib-V-shape} enable point–line correspondences, while orthogonal trihedrons~\cite{Calib-trihedron} and box-like structures~\cite{Calib-Box} provide multiple perpendicular plane constraints for robust extrinsic estimation. These approaches improve geometric observability and reduce ambiguity compared to 2D targets. However, they remain unsuitable for online calibration in dynamic scenes and still rely on reliable target detection in complex environments.
	
	\subsection{Targetless Methods}
	\label{subsec.targetless-methods}
	Targetless calibration methods eliminate the need for specific targets by extracting geometric, semantic, or learned correspondences from natural scenes. Edges of image intensity and LiDAR range are typical geometric correspondences~\cite{Calib-edge1,Calib-edge2,Calib-pixellevel}. Additionally, Neural Radiance Field~\cite{SOAC, MOISST} and 3D Gaussian Splatting~\cite{3DGS-Calib} exploit multi-view geometric and photometric consistency to jointly optimize scene representation and sensor poses. In addition to these geometry-driven approaches, learning-based correspondence methods~\cite{CorrI2P,CoFiI2P} have been developed to align image pixels and LiDAR points within a learned feature space. Other targetless methods rely on objective functions rather than explicit correspondences for extrinsic optimization. For instance, semantic-based methods~\cite{CalibAnything, MIAS-LCEC} maximize the consistency of projected points within segmented image regions. Moreover, information-theoretic methods~\cite{Calib-MI-Grayscale, DirectCalib} estimate the optimal extrinsics by maximizing the mutual information between image grayscale values and projected LiDAR intensity values.

	Beyond these targetless approaches, end-to-end frameworks directly estimate extrinsics without explicit correspondence extraction. Early methods regress extrinsics by fusing RGB images and miscalibrated LiDAR depth maps (e.g., CalibNet~\cite{CalibNet}), sometimes employing VAE-based regularization (RGGNet~\cite{RGGNet}) or monocular depth alignment with LSTM refinement (CalibDepth~\cite{CalibDepth}). Other works construct cost volumes for cross-modal fusion; LCCNet~\cite{LCCNet} computes local feature similarities to build a dense cost volume, which LCCRAFT~\cite{LCCRAFT} iteratively refines using a ConvGRU module. Recently, attention-based methods like MSANet~\cite{MSANet} and CalibFormer~\cite{Calibformer} have emerged, which flatten RGB and depth-map features into tokens to capture global context through cross-attention.

	Our method fundamentally differs from these baselines in its LiDAR encoding space and fusion mechanism. Existing approaches extract features from 2D-projected depth maps, which introduces structural distortion and inevitably discards out-of-frame points under large misalignments---a limitation shared by correlation-based~\cite{CalibNet} and cost-volume-based~\cite{LCCNet} networks. Furthermore, while MSANet~\cite{MSANet} and CalibFormer~\cite{Calibformer} utilize cross-attention for feature matching, they lack a cross-modal coordinate alignment strategy to expand the miscalibrated field of view, severely bottlenecking their interaction capacity under large initial perturbations. In contrast, our extrinsic-aware cross-attention operates directly on native 3D point features, leveraging coordinate alignment to robustly correlate both in-frame and out-of-frame LiDAR geometries with image patches.
	\section{Method}
	\label{sec.method}
	\begin{figure*}[!t]
		\centering
		\includegraphics[width=0.95\linewidth]{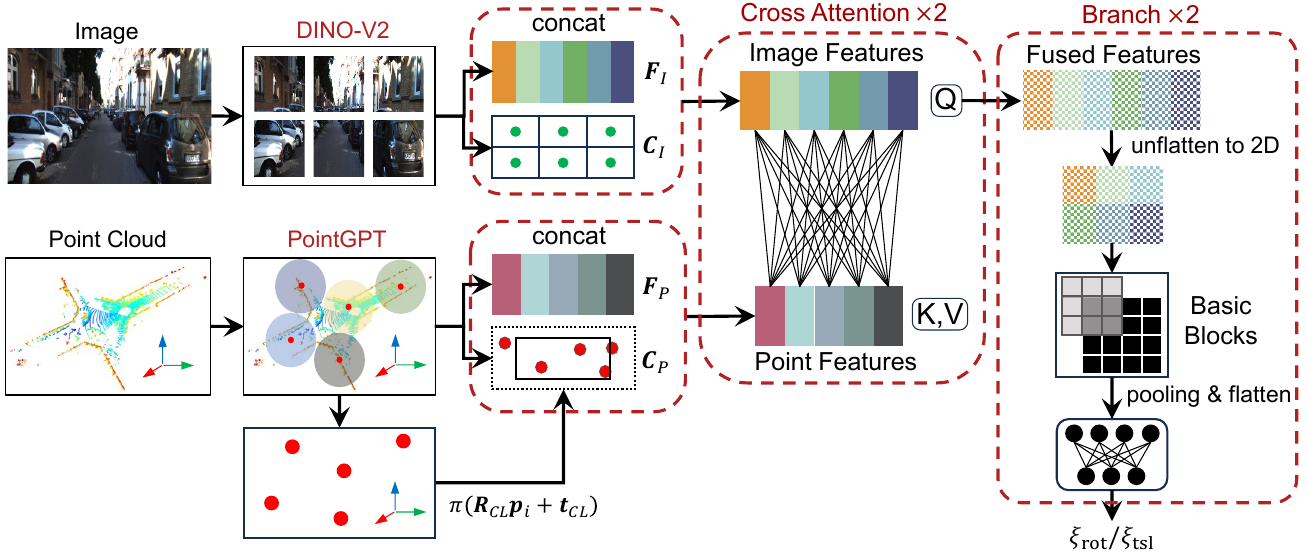}
		\caption{Overall framework of our method. $\bm{F}_I$ and $\bm{F}_P$ denote the sequences of image and point features, respectively, while $\bm{C}_I$ and $\bm{C}_P$ represent their corresponding positional embeddings. $\xi_{\mathrm{rot}}$ and $\xi_{\mathrm{tsl}}$ are the rotational and translational components of $\xi$ defined in \cref{Eq.pred_extrinsic}.}
		\label{Fig.method}
	\end{figure*}
	The overall pipeline of our method is illustrated in~\cref{Fig.method}. The input RGB image and 3D point cloud are first divided into image patches and point groups, respectively, and then encoded into feature vectors. After being concatenated with their corresponding positional embeddings, image and point features are fused through a cross-attention module and subsequently aggregated through convolutional blocks to predict the extrinsic parameters. To decouple rotation and translation learning, the cross-attention and aggregation branches are designed symmetrically for the estimation of rotational and translational components, respectively.
	\subsection{Problem Definition}
	\label{subsec.data_preprocessing}
	Let the RGB image and LiDAR point cloud be denoted as $\bm{I}$ and $\bm{P}$, and the LiDAR and camera coordinate systems as $\bm{O}_L$ and $\bm{O}_C$, respectively.
	Let $\bm{K}\in\mathbb{R}^{3\times3}$ denote the intrinsic camera matrix, and $\bm{T}_{CL}\in\mathbb{R}^{4\times4}$ denote the extrinsic transformation from $\bm{O}_L$ to $\bm{O}_C$.
	
	The intrinsic matrix $\bm{K}$ is typically fixed, whereas the extrinsic matrix $\bm{T}_{CL}$ may vary over time due to factors such as vehicle vibrations or temperature fluctuations.
	If we denote the ground-truth extrinsic matrix as $\bm{T}_{CL}^{gt}$ and the perturbed one after external disturbances as $\bm{T}_{CL}^{(0)}$, the relative perturbation can be expressed as
	$\Delta \bm{T}_{CL} = \bm{T}_{CL}^{gt} (\bm{T}_{CL}^{(0)})^{-1} \in SE(3)$.
	The goal of the calibration model is to estimate $\Delta \bm{T}_{CL}$ given $\bm{I}$, $\bm{P}$, $\bm{K}$, and $\bm{T}_{CL}^{(0)}$ as inputs.
	To remove the geometric constraints of $\Delta \bm{T}_{CL}$, the model predicts its Lie algebra representation $\xi$, which is then mapped back to the Lie group as:
	\begin{equation}
		\label{Eq.pred_extrinsic}
		\bm{T}_{CL}^{pred} = \Delta \bm{T}_{CL}\bm{T}_{CL}^{(0)}= \mathcal{G}(\xi)\bm{T}_{CL}^{(0)},
	\end{equation}
	where $\mathcal{G}(\cdot)$ denotes the exponential mapping from the Lie algebra $\mathfrak{se}(3)$ to the Lie group $SE(3)$, and $\bm{T}_{CL}^{pred}$ represents the predicted extrinsic matrix.
	
	\subsection{Feature Encoding}
	\label{subsec.feature_encoding}
	As illustrated in~\cref{Fig.method}, we adopt DINOv2~\cite{DINOv2}, a pretrained Vision Transformer (ViT)~\cite{ViT}, as the image encoder. The input image is first divided into patches, which are subsequently embedded and encoded into patch-level feature tokens. Additionally, learnable positional embeddings are added to each token to produce spatial awareness. These enriched feature tokens are then processed through cascaded transformer blocks, enabling the model to capture global contextual dependencies across the entire image.
	
	The point encoder follows the architecture of PointGPT~\cite{PointGPT}, where the point cloud is divided into local groups analogous to image patches. Regarding the grouping strategy, a fixed number of centroids are first selected from the original point cloud using Furthest Point Sampling (FPS). Each centroid, along with its k-nearest neighboring points, forms a local group. Each group of points is encoded into a feature vector by PointNet~\cite{PointNet}, and the resulting feature tokens are subsequently processed by transformer layers to capture global geometric context.
	
	Overall, both the RGB image and the LiDAR point cloud are encoded into sequences of feature vectors, denoted as $\bm{F}_I$ and $\bm{F}_P$, whose lengths correspond to the number of image patches and point groups, respectively.
	
	\subsection{Cross-Attention}
	\label{subsec.cross_attention}
	Cross-attention enables interaction between image and point features and produces a fused representation. However, directly cross-attending these modality-specific features is insufficient for estimating the extrinsic matrix: as discussed in \cref{subsec.feature_encoding}, both $\bm{F}_I$ and $\bm{F}_P$ are extracted in their native domains and are inherently independent of the extrinsic matrix $\bm{T}_{CL}$. To address this fundamental limitation, we introduce extrinsic awareness by injecting positional embeddings defined on the image feature plane.
	
	\subsubsection{Image feature plane}
	Let $W$ and $H$ denote the width and height of the original image, and $W_P$ and $H_P$ denote the width and height of each patch, respectively. If each image patch is regarded as a pixel, the set of patches forms an image feature plane $\bm{F}_I^{2D}$ with width $N_W = W / W_P$ and height $N_H = H / H_P$. For numerical stability, patch coordinates are normalized to $[-1,1]$. The coordinate of the patch at the $i$-th row and $j$-th column ($i\in[0,N_H\!-\!1],\, j\in[0,N_W\!-\!1]$) is $\bigl[\,2i/N_H - 1,\; 2j/N_W - 1\,\bigr]$.
	
	\subsubsection{Coordinate Alignment}
	\label{subsubsec.coordinate-alignment}
	To structurally align native 3D LiDAR features with 2D image patches, each LiDAR point $\bm{p}_i$ is first transformed from the LiDAR coordinate system to the camera coordinate system:
	\begin{equation}
		\label{Eq.coord_sys_transform}
		\bm{p}^C_i=\bm{R}_{CL}\bm{p}_i+\bm{t}_{CL},
	\end{equation}
	and then projected onto the image plane:
	\begin{equation}
		\label{Eq.projection}
		w_i\!\begin{bmatrix}\overline{u}_i \\ \overline{v}_i \\ 1\end{bmatrix}
		= \begin{bmatrix}u_i \\ v_i \\ w_i\end{bmatrix}
		= \bm{K}\bm{p}^C_i
		\;\Rightarrow\;
		\begin{bmatrix}\overline{u}_i \\ \overline{v}_i\end{bmatrix}
		= \pi(\bm{p}_i^C),
	\end{equation}
	where $\pi(\cdot)$ denotes the projection operator and $(\overline{u}_i,\overline{v}_i)$ are the pixel coordinates corresponding to $\bm{p}^C_i$. To align the projected LiDAR points with the image patch grid, we scale the projected LiDAR coordinates $(\overline{u}_i,\overline{v}_i)$ by the patch dimensions $(W_P, H_P)$:
	\begin{equation}
		\label{Eq.resized_projection}
		\begin{bmatrix}\tilde{u}_i \\ \tilde{v}_i\end{bmatrix}
		=
		\begin{bmatrix}
			W_P^{-1} & 0 \\
			0 & H_P^{-1}
		\end{bmatrix}
		\pi(\bm{p}_i^C),
	\end{equation}
	and then normalize them to the range $[-1, 1]$ as $(2\tilde{u}_i/N_W - 1, 2\tilde{v}_i/N_H - 1)$. Unlike image patch coordinates, the projected LiDAR coordinates may extend beyond the image region (see \cref{Fig.abstract_depth_proj}). To constrain the maximum projection range, we introduce a margin ratio $r_p$ and clip the normalized coordinates within the range $[-(1+r_p),\,(1+r_p)]$ along both axes to preserve spatial continuity while preventing invalid positions.
	
	\subsubsection{Harmonic embedding}
	\label{subsubsec.harmonic_embedding}
	Inspired by NeRF~\cite{NerF}, we encode 2D coordinates into high-dimensional representations using harmonic functions:
	\begin{equation}
		\label{Eq.harmonic_embedding}
		\begin{aligned}
			\tilde{\bm{x}}_i &= \bigl[\cos(\omega_0 2^0\pi x_i),\,\ldots,\,\cos(\omega_0 2^{n_h-1}\pi x_i),\,x_i\bigr],\\
			\tilde{\bm{y}}_i &= \bigl[\sin(\omega_0 2^0\pi y_i),\,\ldots,\,\sin(\omega_0 2^{n_h-1}\pi y_i),\,y_i\bigr],
		\end{aligned}
	\end{equation}
	where $[x_i,y_i]$ denotes an image or point coordinate pair, and $n_h$ is the number of harmonic functions. The original coordinates are appended to the sinusoidal embeddings to retain absolute positional information and complement the multi-frequency encoding. This representation introduces positional cues at multiple frequencies, enhancing the cross-attention module’s sensitivity to fine-grained spatial relationships. To ensure that the longest period of the harmonic embedding precisely covers the marginal range $[-(1+r_p),\,(1+r_p)]$, we set $\omega_0 = 1/(1+r_p)$.
	
	\subsubsection{Multi-head attention}
	Denote $[\cdot\,;\,\cdot]$ as channel-wise concatenation. Then, stacking the pairs $[\tilde{\bm{x}}_i;\tilde{\bm{y}}_i]$ yields the positional embeddings derived from the image-plane and projected LiDAR coordinates, denoted as $\bm{C}_I$ and $\bm{C}_P$, respectively. We \emph{concatenate} positional embeddings and features along the channel dimension to form the tokens:
	\begin{equation}
		\label{Eq.feature_coord_concat}
		\begin{aligned}
			\bm{X} &= [\,\bm{F}_I \,;\, \bm{C}_I\,], \\
			\bm{Y} &= [\,\bm{F}_P \,;\, \bm{C}_P\,]
		\end{aligned}
	\end{equation}
	Following improvements from prior ViT literature~\cite{NaViT}, we adopt a scale-free cross-attention variant:
	\begin{align}
		\label{Eq.q_norm}
		\bm{Q}_i&=\mathrm{RMSNorm}(\mathrm{LayerNorm}(\bm{X})\bm{W}_i^Q)\\
		\label{Eq.k_norm}
		\bm{K}_i&=\mathrm{RMSNorm}(\bm{X}\bm{W}_i^K),\,\bm{V}_i=\bm{Y}\bm{W}_i^V\\
		\label{Eq.scaled_product}
		\bm{A}_i&=\mathrm{Softmax}(\bm{Q}_i\bm{K}_i^\top)\bm{V}_i \\
		\label{Eq.out_linear}
		\bm{O}&=[\bm{A}_1;\bm{A}_2;\ldots;\bm{A}_h]\bm{W}^O
	\end{align}
	where $\bm{W}_i^Q,\bm{W}_i^K,\bm{W}_i^V,\bm{W}_i^O$ are projection weights, $h$ is the number of heads, and $\bm{O}$ is the module output. Since rotation and translation estimation benefit from distinct cues, we duplicate the cross-attention block and output two feature vectors, which are then processed separately for rotational and translational components.
	
	\subsection{Aggregation}
	\label{subsec.aggregation}
	Following prior works~\cite{CalibNet, LCCNet, RGGNet}, our method employs convolutional kernels for feature aggregation and MLP layers for extrinsic regression. While previous approaches use only separate linear projections for rotation and translation estimation, we hypothesize that earlier modules—such as convolutional blocks—should also be decoupled to enable finer-grained feature aggregation. Since the cross-attention modules output 1D feature vectors, we first reshape (unflatten) them into 2D feature maps with the same spatial resolution as $\bm{F}_I^{2D}$. We then apply two independent basic blocks~\cite{Resnet} for feature encoding of the rotational and translational branches, respectively. The resulting feature maps are spatially pooled into compact representations and subsequently flattened into 1D vectors for MLP regression. Finally, two separate MLP heads are used to predict $\xi_{\mathrm{rot}}$ and $\xi_{\mathrm{tsl}}$, \ie, the rotational and translational components of the extrinsic update $\xi\in \mathfrak{se}(3)$.
	
	\section{Experiments}
	\label{sec.experiments}
	\subsection{Dataset Description}
	\label{subsec.dataset}
	We evaluate our method against state-of-the-art learning-based~\cite{CalibNet,RGGNet,LCCNet,LCCRAFT,CalibDepth,CoFiI2P} and learning-free approaches~\cite{DirectCalib,CalibAnything} on the KITTI Odometry~\cite{KITTI} and nuScenes~\cite{nuScenes} datasets. As implementations for recent cross-attention baselines~\cite{MSANet,Calibformer} are publicly unavailable, we design a targeted ablation study in \cref{subsec.ablation_analysis}. This experiment simulates their core mechanism to explicitly demonstrate the superiority of our cross-modal coordinate alignment and depth map expansion under severe miscalibration.

	For dataset splits, we use KITTI sequences 00, 02--08, 10, 12, and 21 for training, 11, 17, and 20 for validation, and 13--16 and 18 for testing. For nuScenes, we follow the official split and reserve 20\% of the training data for validation. 

	These two datasets pose distinct challenges: nuScenes contains sparser LiDAR point clouds and more nighttime scenes, while KITTI exhibits a more severe train--test distribution shift. Specifically, the training--testing gap measured by Fr\'echet Inception Distance (FID) is 39.60 for KITTI compared to 15.22 for nuScenes, indicating stronger generalization demands on the KITTI dataset.
	
	To evaluate calibration performance under different initialization errors, we perturb the ground-truth extrinsics to generate the initial extrinsic matrix:
	\[
	\bm{T}_{CL}^{(0)} = \bm{T}_r \bm{T}_{CL}^{gt},
	\]
	where $\bm{T}_r$ introduces rotational and translational perturbations of \sepdegcm{15}{15}, \sepdegcm{10}{25}, and \sepdegcm{10}{50}.

	\begin{table*}[!t]
		\centering
		\caption{Calibration Results on KITTI and nuScenes Datasets (Mean $\pm$ Standard Deviation)}
		\label{Table.calib_metrics}
		\resizebox{\textwidth}{!}{
			\begin{tabular}{lllcc|cc|cc}
				\toprule
				\multirow{2}*{Dataset} & \multirow{2}*{Range} & \multirow{2}*{Method} & \multicolumn{2}{c}{Rotation ($^\circ$)$\downarrow$} & \multicolumn{2}{c}{Translation (cm)$\downarrow$} & \multicolumn{2}{c}{Success Rate (\%)$\uparrow$} \\
				~ & ~ & ~ & RMSE & MAE & RMSE & MAE & $L_1$ & $L_2$ \\
				\midrule
				\multirow{27}*{KITTI~\cite{KITTI}} & \multirow{9}*{$15^\circ\,15\mathrm{cm}$} & CoFiI2P\cite{CoFiI2P} & 4.613$\pm$3.071 & 2.066$\pm$1.228 & 134.8$\pm$75.09 & 62.64$\pm$32.60 & 0.00\% & 0.04\% \\
				~ & ~ & DirectCalib\cite{DirectCalib} & 13.09$\pm$23.55 & 6.315$\pm$11.38 & 194.9$\pm$1967 & 98.42$\pm$1099 & 0.26\% & 1.54\% \\
				~ & ~ & CalibAnything\cite{CalibAnything} & 18.27$\pm$14.78 & 9.439$\pm$7.889 & 27.25$\pm$15.11 & 13.79$\pm$7.742 & 0.00\% & 1.90\% \\
				~ & ~ & CalibNet\cite{CalibNet} & 2.019$\pm$2.104 & 0.764$\pm$0.726 & 5.798$\pm$3.598 & 2.836$\pm$1.783 & 8.00\% & 32.28\% \\
				~ & ~ & RGGNet\cite{RGGNet} & 3.878$\pm$3.380 & 1.421$\pm$1.199 & 6.069$\pm$4.037 & 2.971$\pm$2.016 & 5.40\% & 18.56\% \\
				~ & ~ & LCCNet\cite{LCCNet} & 2.095$\pm$2.208 & 0.804$\pm$0.784 & 6.121$\pm$4.076 & 3.012$\pm$2.055 & 9.16\% & 31.72\% \\
				~ & ~ & LCCRAFT\cite{LCCRAFT} & \underline{0.530$\pm$0.784} & \textbf{0.206$\pm$0.270} & 6.030$\pm$3.553 & 2.897$\pm$1.702 & 11.20\% & 44.12\% \\
				~ & ~ & CalibDepth\cite{CalibDepth} & 1.057$\pm$1.199 & 0.418$\pm$0.414 & \underline{4.573$\pm$2.798} & \underline{2.230$\pm$1.344} & \underline{17.24\%} & \underline{56.88\%} \\
				~ & ~ & Ours & \textbf{0.431$\pm$1.045} & \underline{0.212$\pm$0.500} & \textbf{2.199$\pm$1.816} & \textbf{1.087$\pm$0.902} & \textbf{54.64\%} & \textbf{96.64\%} \\
				\cdashline{3-9}[1pt/1pt]
				~ & \multirow{9}*{$10^\circ\,25\mathrm{cm}$} & CoFiI2P & 2.939$\pm$2.134 & 1.285$\pm$0.841 & 60.74$\pm$33.26 & 28.08$\pm$14.59 & 0.04\% & 0.12\% \\
				~ & ~ & DirectCalib & 13.08$\pm$26.34 & 6.464$\pm$13.15 & 147.1$\pm$401.2 & 69.93$\pm$190.9 & 0.38\% & 1.79\% \\
				~ & ~ & CalibAnything & 5.323$\pm$8.955 & 2.598$\pm$4.530 & 28.20$\pm$25.50 & 14.16$\pm$13.08 & 0.95\% & 12.38\% \\
				~ & ~ & CalibNet & 2.280$\pm$2.379 & 0.891$\pm$0.908 & 6.466$\pm$3.746 & 3.151$\pm$1.842 & 4.24\% & 26.64\% \\
				~ & ~ & RGGNet & 3.987$\pm$3.492 & 1.524$\pm$1.391 & 6.235$\pm$4.088 & 3.045$\pm$2.025 & 4.88\% & 17.84\% \\
				~ & ~ & LCCNet & 2.496$\pm$2.532 & 0.959$\pm$0.946 & 6.083$\pm$3.867 & 2.978$\pm$1.933 & 7.16\% & 27.76\% \\
				~ & ~ & LCCRAFT & \textbf{0.593$\pm$0.783} & \textbf{0.229$\pm$0.269} & 6.271$\pm$3.906 & 2.949$\pm$1.792 & \underline{11.20\%} & \underline{42.44\%} \\
				~ & ~ & CalibDepth & 1.989$\pm$2.479 & 0.744$\pm$0.894 & \underline{5.441$\pm$3.389} & \underline{2.597$\pm$1.623} & 9.40\% & 39.16\% \\
				~ & ~ & Ours & \underline{0.654$\pm$1.435} & \underline{0.319$\pm$0.646} & \textbf{2.594$\pm$1.755} & \textbf{1.292$\pm$0.888} & \textbf{48.84\%} & \textbf{92.56\%} \\
				\cdashline{3-9}[1pt/1pt]
				~ & \multirow{9}*{$10^\circ\,50\mathrm{cm}$} & CoFiI2P & 2.897$\pm$2.182 & 1.263$\pm$0.860 & 87.00$\pm$38.02 & 38.33$\pm$15.83 & 0.00\% & 0.00\% \\
				~ & ~ & DirectCalib & 12.66$\pm$24.06 & 6.216$\pm$11.71 & 223.0$\pm$1394 & 110.0$\pm$728.9 & 0.00\% & 0.77\% \\
				~ & ~ & CalibAnything & 6.024$\pm$9.581 & 2.900$\pm$4.737 & 49.91$\pm$48.21 & 24.90$\pm$24.36 & 0.95\% & 8.57\% \\
				~ & ~ & CalibNet & 2.339$\pm$2.388 & 0.925$\pm$0.914 & 8.304$\pm$4.912 & 4.028$\pm$2.385 & 2.04\% & 17.36\% \\
				~ & ~ & RGGNet & 4.032$\pm$3.533 & 1.570$\pm$1.437 & 6.505$\pm$4.065 & 3.183$\pm$2.019 & 4.08\% & 16.44\% \\
				~ & ~ & LCCNet & 2.548$\pm$2.551 & 0.994$\pm$0.958 & 6.723$\pm$4.550 & 3.286$\pm$2.254 & 6.04\% & 25.56\% \\
				~ & ~ & LCCRAFT & \underline{0.951$\pm$1.117} & \textbf{0.352$\pm$0.386} & 6.485$\pm$4.199 & 3.084$\pm$2.067 & \underline{9.16\%} & 39.08\% \\
				~ & ~ & CalibDepth & 1.775$\pm$2.143 & 0.668$\pm$0.738 & \underline{5.275$\pm$3.200} & \underline{2.557$\pm$1.520} & 8.68\% & \underline{41.76\%} \\
				~ & ~ & Ours & \textbf{0.764$\pm$0.911} & \underline{0.371$\pm$0.436} & \textbf{2.747$\pm$1.427} & \textbf{1.363$\pm$0.705} & \textbf{41.04\%} & \textbf{87.68\%} \\
				\midrule
				\multirow{27}*{nuScenes~\cite{nuScenes}} & \multirow{9}*{$15^\circ\,15\mathrm{cm}$} & CoFiI2P & 5.085$\pm$4.312 & 2.504$\pm$1.957 & 179.2$\pm$97.26 & 81.67$\pm$46.58 & 0.00\% & 0.00\% \\
				~ & ~ & DirectCalib & 14.89$\pm$22.03 & 7.182$\pm$10.29 & 451.8$\pm$1633 & 212.4$\pm$773.7 & 0.00\% & 0.17\% \\
				~ & ~ & CalibAnything & 7.512$\pm$4.565 & 3.895$\pm$2.480 & 7.240$\pm$4.983 & 3.773$\pm$2.711 & 0.52\% & 3.30\% \\
				~ & ~ & CalibNet & 2.121$\pm$1.997 & 0.896$\pm$0.844 & 6.335$\pm$3.827 & 2.900$\pm$1.716 & 8.24\% & 35.04\% \\
				~ & ~ & RGGNet & 4.205$\pm$3.504 & 1.756$\pm$1.504 & 6.063$\pm$4.090 & 2.879$\pm$1.952 & 4.61\% & 17.35\% \\
				~ & ~ & LCCNet & 2.344$\pm$2.469 & 1.005$\pm$1.100 & 5.588$\pm$4.489 & 2.642$\pm$2.211 & 13.65\% & 41.91\% \\
				~ & ~ & LCCRAFT & 0.708$\pm$1.648 & 0.275$\pm$0.573 & 5.570$\pm$4.481 & 2.286$\pm$1.704 & 27.09\% & 57.52\% \\
				~ & ~ & CalibDepth & \textbf{0.299$\pm$0.196} & \textbf{0.150$\pm$0.103} & \underline{3.326$\pm$2.637} & \underline{1.381$\pm$0.955} & \underline{48.90\%} & \underline{79.14\%} \\
				~ & ~ & Ours & \underline{0.366$\pm$0.228} & \underline{0.185$\pm$0.119} & \textbf{0.506$\pm$0.278} & \textbf{0.246$\pm$0.134} & \textbf{97.89\%} & \textbf{99.93\%} \\
				\cdashline{3-9}[1pt/1pt]
				~ & \multirow{9}*{$10^\circ\,25\mathrm{cm}$} & CoFiI2P & 3.843$\pm$2.151 & 1.863$\pm$1.026 & 104.6$\pm$79.45 & 49.77$\pm$37.49 & 0.00\% & 0.00\% \\
				~ & ~ & DirectCalib & 13.25$\pm$22.32 & 6.225$\pm$10.25 & 267.6$\pm$1005 & 122.7$\pm$463.4 & 0.17\% & 0.17\% \\
				~ & ~ & CalibAnything & 4.734$\pm$3.168 & 2.477$\pm$1.720 & 11.94$\pm$8.318 & 6.253$\pm$4.504 & 1.74\% & 5.73\% \\
				~ & ~ & CalibNet & 2.098$\pm$1.976 & 0.888$\pm$0.836 & 6.336$\pm$3.840 & 2.897$\pm$1.720 & 8.37\% & 35.86\% \\
				~ & ~ & RGGNet & 3.949$\pm$3.327 & 1.558$\pm$1.296 & 6.028$\pm$4.088 & 2.853$\pm$1.942 & 5.21\% & 18.30\% \\
				~ & ~ & LCCNet & 2.406$\pm$2.559 & 1.037$\pm$1.143 & 5.858$\pm$5.719 & 2.781$\pm$2.873 & 13.55\% & 41.54\% \\
				~ & ~ & LCCRAFT & 0.636$\pm$1.206 & 0.251$\pm$0.421 & 5.629$\pm$4.276 & 2.308$\pm$1.603 & 24.78\% & 55.01\% \\
				~ & ~ & CalibDepth & \underline{0.393$\pm$0.251} & \underline{0.196$\pm$0.127} & \underline{3.778$\pm$2.844} & \underline{1.583$\pm$1.039} & \underline{41.26\%} & \underline{74.10\%} \\
				~ & ~ & Ours & \textbf{0.392$\pm$0.281} & \textbf{0.194$\pm$0.145} & \textbf{0.526$\pm$0.330} & \textbf{0.257$\pm$0.160} & \textbf{97.18\%} & \textbf{99.70\%} \\
				\cdashline{3-9}[1pt/1pt]
				~ & \multirow{9}*{$10^\circ\,50\mathrm{cm}$} & CoFiI2P & 3.743$\pm$2.327 & 1.766$\pm$1.051 & 54.16$\pm$33.93 & 26.31$\pm$16.86 & 0.00\% & 0.00\% \\
				~ & ~ & DirectCalib & 12.74$\pm$22.00 & 6.209$\pm$10.68 & 359.0$\pm$1156 & 169.0$\pm$544.8 & 0.00\% & 0.33\% \\
				~ & ~ & CalibAnything & 4.734$\pm$3.168 & 2.477$\pm$1.720 & 23.88$\pm$16.64 & 12.51$\pm$9.008 & 0.52\% & 3.47\% \\
				~ & ~ & CalibNet & 2.470$\pm$2.284 & 1.023$\pm$0.912 & 8.783$\pm$5.665 & 3.987$\pm$2.561 & 4.01\% & 20.94\% \\
				~ & ~ & RGGNet & 5.827$\pm$4.176 & 2.707$\pm$2.029 & 7.250$\pm$4.832 & 3.771$\pm$2.617 & 2.32\% & 8.47\% \\
				~ & ~ & LCCNet & 2.829$\pm$3.150 & 1.212$\pm$1.472 & 7.559$\pm$12.08 & 3.596$\pm$6.163 & 6.97\% & 29.41\% \\
				~ & ~ & LCCRAFT & 0.937$\pm$1.539 & 0.365$\pm$0.534 & 7.407$\pm$5.334 & 3.203$\pm$2.221 & 13.72\% & 40.14\% \\
				~ & ~ & CalibDepth & \textbf{0.363$\pm$0.236} & \textbf{0.182$\pm$0.120} & \underline{5.711$\pm$4.593} & \underline{2.222$\pm$1.610} & \underline{27.77\%} & \underline{54.38\%} \\
				~ & ~ & Ours & \underline{0.595$\pm$0.364} & \underline{0.299$\pm$0.187} & \textbf{0.775$\pm$0.459} & \textbf{0.382$\pm$0.223} & \textbf{89.81\%} & \textbf{99.16\%} \\
				\bottomrule
			\end{tabular}
		}
	\end{table*}
	\subsection{Implementation Details}
	\label{subsec.implementation}
	We employ DINOv2-tiny~\cite{DINOv2} as the image encoder and PointGPT-tiny~\cite{PointGPT} as the point encoder, each producing feature embeddings of 384 channels. For the positional embedding, we use six harmonic functions and set the projection margin to $r_p=2$. The multi-head cross-attention module contains 6 heads, each with a 64-dimensional subspace. The aggregation module consists of two basic residual blocks~\cite{Resnet}, which progressively reduce the feature dimensionality from 384 to 96.
	Within the MLP layers, the hidden dimension is set to 128, and SiLU~\cite{SiLU} is adopted as the activation function.
	
	Regarding training configurations, the input point clouds are downsampled to 40,000 points for the KITTI Odometry dataset and 20,000 points for the nuScenes dataset. Input images are resized to $224\times448$ ($H\times W$) for both datasets. During inference, we adopt a three-step iterative refinement strategy: the predicted extrinsic matrix from each step is used as the initial estimate for the next iteration, and the final prediction after three iterations is taken as the output.
	
	\subsection{Metrics}
	\label{subsec.metrics}
	We evaluate the calibration accuracy by computing the pose difference between the predicted and ground-truth extrinsic matrices,
	\ie, $\Delta \bm{T}_{\mathrm{err}} = \bm{T}_{CL}^{\mathrm{pred}} \bigl(\bm{T}_{CL}^{\mathrm{gt}}\bigr)^{-1}$. From $\Delta \bm{T}_{\mathrm{err}}$, we extract the Euler angles (Roll, Pitch, Yaw) representing rotational errors and the translation components (X, Y, Z) representing translational errors, and compute the root mean squared error (RMSE) for both rotation and translation.
	
	In addition, we assess calibration robustness using two success-rate metrics, denoted as $L_1$ and $L_2$. Specifically, $L_1$ measures the percentage of predictions whose rotational RMSE is below $1^\circ$ and translational RMSE is below 2.5 cm, while $L_2$ adopts a more relaxed threshold of \sepdegcm{2}{5}.
	
	\subsection{Evaluation}
	\label{subsec.evaluation}
	\begin{figure*}[!t]
		\includegraphics[width=0.95\linewidth]{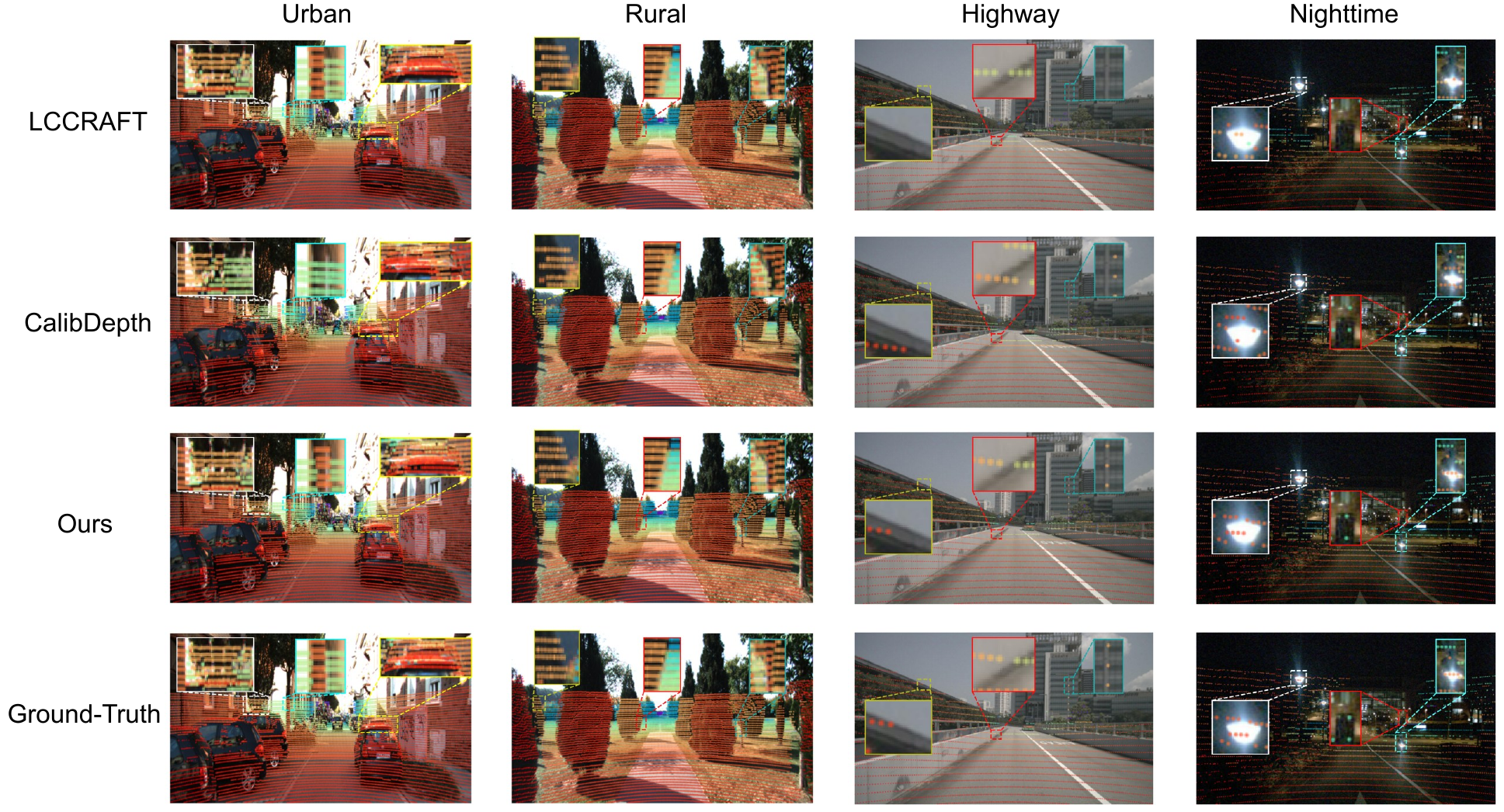}
		\caption{
			LiDAR projection maps generated using the predicted extrinsic matrix $\bm{T}_{CL}^{\mathrm{pred}}$ from different methods across urban, rural, highway, and nighttime scenes. Selected regions of interest (ROIs) are zoomed in for clearer visualization.
		}
		\label{Fig.res_cmp}
	\end{figure*}
	~\Cref{Table.calib_metrics} presents the quantitative calibration results on the KITTI~\cite{KITTI} and nuScenes~\cite{nuScenes} datasets. On the KITTI Odometry dataset, our proposed method consistently achieves superior performance over baselines across most metrics under all initialization ranges, except for a slightly higher rotational RMSE than LCCRAFT under \degcm{10}{25}. Notably, our approach demonstrates a clear advantage in translation accuracy, achieving a translational RMSE nearly half that of the closest competitor across all initialization settings. Although the performance of all methods degrades as the initial translation error increases, our method remains the most robust, achieving 88\% $L_2$ success rate and 41\% $L_1$ success rate under the most challenging perturbation of \degcm{10}{50}. 
	
	Compared with KITTI, nuScenes has a smaller train–test distribution gap but uses a 32-beam LiDAR instead of 64 beams. Our method achieves lower translation errors and higher success rates than the baselines on nuScenes. Although its rotational errors are slightly higher than CalibDepth under \degcm{15}{15} and \degcm{10}{50}, the differences are marginal and do not affect overall success rates. While learning-based baselines perform relatively better on nuScenes under \degcm{15}{15} and \degcm{10}{25}, their performance degrades under the challenging setting of \degcm{10}{50}. In contrast, our method remains robust, achieving 99\% $L_2$ and 90\% $L_1$ success rates.
	
	Learning-free methods perform relatively poorly in our experiments. DirectCalib~\cite{DirectCalib} optimizes extrinsics by maximizing mutual information and relies on dense projected LiDAR intensity maps, which are more suitable for static scanning scenarios and differ from the sparse and dynamic autonomous driving data used in our experiments. CalibAnything~\cite{CalibAnything} is primarily designed for small rotational perturbations and thus degrades significantly under larger rotations.
	
	We also present qualitative comparisons with two strongest baselins, LCCRAFT and CalibDepth, across urban, rural, highway, and nighttime scenes (\cref{Fig.res_cmp}). ROI zoom-ins show that our method achieves more consistent alignment across environments, particularly along object boundaries such as vehicle contours, tree trunks, guardrails, and headlights. In contrast, LCCRAFT aligns well in urban and rural scenes but degrades in highway and nighttime conditions, whereas CalibDepth performs better in highway and nighttime scenes but is less accurate in other scenarios.
	
	This robustness stems from incorporating LiDAR point features projected beyond the image frame. As indicated by \cref{Eq.coord_sys_transform} and \cref{Eq.projection}, when the extrinsics $\bm{R}_{CL}$ and $\bm{t}_{CL}$ deviate significantly from the ground truth, projected LiDAR points may shift substantially or fall outside the image plane. In such cases (\cref{Fig.abstract_depth_proj}), simple concatenation of image features and miscalibrated depth maps fails to establish reliable correspondences, whereas our cross-attention mechanism preserves and exploits these cross-modal relationships.
	
	\subsection{Ablation Analysis}
	\label{subsec.ablation_analysis}
	\begin{table*}[htbp]
		\centering
		\caption{Ablation on KITTI at $10^\circ\,50\mathrm{cm}$ (Mean $\pm$ Standard Deviation)}
		\label{Table.ablation}
		\resizebox{\textwidth}{!}{
			\begin{tabular}{cccccccccc}
				\toprule
				\multirow{2}{*}{Index} & \multirow{2}{*}{\makecell{Dual \\ Branches}} & \multirow{2}{*}{\makecell{Projection \\ Margin}} & \multirow{2}{*}{\makecell{Encoding \\ Space}} & \multirow{2}{*}{\makecell{Positional \\ Embedding}} & \multirow{2}{*}{\makecell{Image \\ Encoder}} & \multirow{2}{*}{\makecell{Rotation \\ RMSE ($^\circ$)$\downarrow$}} & \multirow{2}{*}{\makecell{Translation \\ RMSE (cm)$\downarrow$}} & \multicolumn{2}{c}{Success Rate (\%)$\uparrow$} \\
				& & & & & & & & $L_1$ & $L_2$ \\
				\midrule
				1 & \cmark & \xmark & 3D & concatenation & DINOv2 & 1.534$\pm$1.304 & 3.002$\pm$2.861 & 22.60\% & 73.64\% \\
				2 & \cmark & \cmark & 2D & harmonic ($n_h=6$) & DINOv2 & \textbf{0.244$\pm$0.279} & 3.429$\pm$2.490 & 39.84\% & 83.12\% \\
				3 & \cmark & \xmark & 2D & harmonic ($n_h=6$) & DINOv2 & 1.111$\pm$1.055 & 15.56$\pm$8.714 & 0.60\% & 6.12\% \\
				4 & \cmark & \cmark & 3D & harmonic ($n_h=0$) & DINOv2 & 1.297$\pm$1.197 & 3.199$\pm$1.591 & 20.12\% & 76.88\% \\
				5 & \cmark & \cmark & 3D & harmonic ($n_h=2$) & DINOv2 & 0.728$\pm$0.893 & \underline{2.845$\pm$3.767} & \underline{40.76\%} & \textbf{89.16\%} \\
				\rowcolor{gray!15}
				6 &\cmark & \cmark & 3D & harmonic ($n_h=6$) & DINOv2 & 0.764$\pm$0.911 & \textbf{2.747$\pm$1.427} & \textbf{41.04\%} & \underline{87.68\%} \\
				7 & \cmark & \cmark & 3D & harmonic ($n_h=10$) & DINOv2 & 0.825$\pm$1.134 & 2.924$\pm$1.733 & 37.24\% & 85.52\% \\
				8 & \cmark & \xmark & 3D & harmonic ($n_h=6$) & DINOv2 & 1.407$\pm$1.923 & 8.786$\pm$21.41 & 25.08\% & 73.92\% \\
				9 & \cmark & \cmark & 3D & RoPE-2D ($f_B=10^3$) & DINOv2 & 0.786$\pm$0.812 & 3.251$\pm$3.888 & 32.20\% & 85.40\% \\
				10 & \xmark & \cmark & 3D & harmonic ($n_h=6$) & DINOv2 & 1.019$\pm$1.101 & 3.031$\pm$2.465 & 33.88\% & 84.32\% \\
				11 & \cmark & \cmark & 3D & harmonic ($n_h=6$) & ResNet-18 & \underline{0.726$\pm$1.033} & 2.994$\pm$1.381 & 36.08\% & 85.68\% \\
				\bottomrule
			\end{tabular}
		}
	\end{table*}
	We ablate key components of our method on KITTI under \degcm{10}{50} in \cref{Table.ablation}, where the default configuration is highlighted in light gray. Replacing cross-attention with feature concatenation in 2D ($1^\mathrm{st}$ row) degrades all metrics. Using a 2D depth-map encoder instead of our 3D encoder ($2^\mathrm{nd}$ row) reduces translation accuracy and success rates, although rotation improves, possibly due to the enlarged projection margin compensating for 2D spatial distortions. 

	For positional encoding, removing the harmonic formulation ($n_h=0$) leads to a substantial performance drop. Setting $n_h=6$ achieves the best trade-off between translation accuracy and success rates, consistently outperforming smaller $n_h$ variants and RoPE-2D~\cite{RoPE} (with base frequency $f_B=1000$).

	Crucially, we validate the effectiveness of field-of-view expansion by masking out-of-frame points (i.e., removing the projection margin). As shown by the $6^\mathrm{th}$ vs. $8^\mathrm{th}$ rows in the 3D encoding space and the $2^\mathrm{nd}$ vs. $3^\mathrm{rd}$ rows in the 2D encoding space, removing this margin leads to a significant performance degradation. In the 2D setting, the translation RMSE increases from 3.429\,cm to 15.56\,cm, while the $L_1$ success rate drops from 39.84\% to 0.60\%. These results highlight the importance of preserving out-of-frame geometries for robust cross-attention under large initial perturbations, distinguishing our design from prior attention-based methods such as CalibFormer~\cite{Calibformer} and MSANet~\cite{MSANet}.

	Additionally, decoupling the aggregation branches for $\xi_\mathrm{rot}$ and $\xi_\mathrm{tsl}$ ($6^\mathrm{th}$ vs. $10^\mathrm{th}$ row) reduces rotational and translational RMSE by 25.0\% and 9.4\%, respectively. Finally, substituting the DINOv2~\cite{DINOv2} image encoder with ResNet-18~\cite{Resnet} ($6^\mathrm{th}$ vs. $11^\mathrm{th}$ row) drops overall success rates by 3\%--5\%, demonstrating DINOv2's superior complementary representations for cross-modal calibration.
	
	\section{Conclusion}
	
	In this paper, we analyze the limitations of existing camera–LiDAR calibration methods that rely on miscalibrated depth maps and propose an extrinsic-aware cross-attention framework to address these issues. Extensive experiments on the KITTI and nuScenes datasets validate the effectiveness of our method, achieving improved accuracy and robustness compared with state-of-the-art baselines.
	
	For future work, we plan to extend cross-attention beyond semantic features to incorporate structural cues such as lines, edges, and object-level geometry, further enhancing interpretability and reliability. Another promising direction is to expand the image feature plane to better accommodate the large spatial extent of projected depth maps, enabling more stable calibration under extreme misalignment.

	\bibliographystyle{IEEEtran}
	\bibliography{ref}\ 

@inproceedings{Calib-chessboard-line-plane,
  title={Automatic extrinsic calibration of a camera and a 3d lidar using line and plane correspondences},
  author={Zhou, Lipu and Li, Zimo and Kaess, Michael},
  booktitle={IROS},
  pages={5562--5569},
  year={2018},
  organization={IEEE}
}

@article{Calib-unni-chessboard,
  title={Fast extrinsic calibration of a laser rangefinder to a camera},
  author={Unnikrishnan, Ranjith and Hebert, Martial},
  journal={Robotics Institute, Pittsburgh, PA, Tech. Rep. CMU-RI-TR-05-09},
  year={2005}
}

@article{Calib-chessboard-plane-only,
  title={Extrinsic calibration of a 3d laser scanner and an omnidirectional camera},
  author={Pandey, Gaurav and McBride, James and Savarese, Silvio and Eustice, Ryan},
  journal={IFAC Proceedings Volumes},
  volume={43},
  number={16},
  pages={336--341},
  year={2010},
  publisher={Elsevier}
}

@inproceedings{Calib-checkboard-passthrough,
  title={A Novel, Efficient and Accurate Method for Lidar Camera Calibration},
  author={Huang, Zhanhong and Zhang, Xiao and Garcia, Antony and Huang, Xinming},
  booktitle={ICRA},
  pages={14513--14519},
  year={2024},
  organization={IEEE}
}

@InProceedings{Calib-Box,
author = {Pusztai, Zoltan and Hajder, Levente},
title = {Accurate Calibration of LiDAR-Camera Systems Using Ordinary Boxes},
booktitle = {ICCV Workshops},
month = {Oct},
year = {2017},
doi={10.1109/ICCVW.2017.53}
}

@article{Calib-trihedron,
  title={Extrinsic calibration of a 3D LIDAR and a camera using a trihedron},
  author={Gong, Xiaojin and Lin, Ying and Liu, Jilin},
  journal={Optics and Lasers in Engineering},
  volume={51},
  number={4},
  pages={394--401},
  year={2013},
  publisher={Elsevier}
}

@inproceedings{Calib-V-shape,
  title={Extrinsic calibration of a single line scanning lidar and a camera},
  author={Kwak, Kiho and Huber, Daniel F and Badino, Hernan and Kanade, Takeo},
  booktitle={IROS},
  pages={3283--3289},
  year={2011},
  organization={IEEE}
}

@article{Calib-board,
  title={An effective camera-to-LiDAR spatiotemporal calibration based on a simple calibration target},
  author={Grammatikopoulos, Lazaros and Papanagnou, Anastasios and Venianakis, Antonios and Kalisperakis, Ilias and Stentoumis, Christos},
  journal={Sensors},
  volume={22},
  number={15},
  pages={5576},
  year={2022},
  publisher={MDPI}
}

@inproceedings{Calib-Cycle-Hole,
  title={Automatic extrinsic calibration for lidar-stereo vehicle sensor setups},
  author={Guindel, Carlos and Beltr{\'a}n, Jorge and Mart{\'\i}n, David and Garc{\'\i}a, Fernando},
  booktitle={ITSC},
  pages={1--6},
  year={2017},
  organization={IEEE}
}

@article{Calib-triangular,
title={Calibration between color camera and 3D LIDAR instruments with a polygonal planar board},
author={Park, Yoonsu and Yun, Seokmin and Won, Chee Sun and Cho, Kyungeun and Um, Kyhyun and Sim, Sungdae},
journal={Sensors},
volume={14},
number={3},
pages={5333--5353},
year={2014},
publisher={MDPI}
}

@article{Calib-triangular2,
  title={LiDAR--camera calibration method based on ranging statistical characteristics and improved RANSAC algorithm},
  author={Xu, Xiaobin and Zhang, Lei and Yang, Jian and Liu, Cong and Xiong, Yiyang and Luo, Minzhou and Tan, Zhiying and Liu, Bo},
  journal={Robotics and Autonomous Systems},
  volume={141},
  pages={103776},
  year={2021},
  publisher={Elsevier}
}

@inproceedings{L2v2t2,
  title={L 2 v 2 t 2 calib: Automatic and unified extrinsic calibration toolbox for different 3d lidar, visual camera and thermal camera},
  author={Zhang, Jun and Liu, Yiyao and Wen, Mingxing and Yue, Yufeng and Zhang, Haoyuan and Wang, Danwei},
  booktitle={2023 IEEE Intelligent Vehicles Symposium (IV)},
  pages={1--7},
  year={2023},
  organization={IEEE}
}

@article{Object-Detection1,
  title={LiDAR-Camera Fusion in Perspective View for 3D Object Detection in Surface Mine},
  author={Ai, Yunfeng and Yang, Xue and Song, Ruiqi and Cui, Chenglin and Li, Xinqing and Cheng, Qi and Tian, Bin and Chen, Long},
  journal={IEEE Transactions on Intelligent Vehicles},
  year={2023},
  publisher={IEEE}
}

@inproceedings{Object-Detection2,
  title={Virtual sparse convolution for multimodal 3d object detection},
  author={Wu, Hai and Wen, Chenglu and Shi, Shaoshuai and Li, Xin and Wang, Cheng},
  booktitle={CVPR},
  pages={21653--21662},
  year={2023}
}

@inproceedings{SLAM1,
  title={R 3 LIVE: A Robust, Real-time, RGB-colored, LiDAR-Inertial-Visual tightly-coupled state Estimation and mapping package},
  author={Lin, Jiarong and Zhang, Fu},
  booktitle={ICRA},
  pages={10672--10678},
  year={2022},
  organization={IEEE}
}

@inproceedings{SLAM2,
  title={Camvox: A low-cost and accurate lidar-assisted visual slam system},
  author={Zhu, Yuewen and Zheng, Chunran and Yuan, Chongjian and Huang, Xu and Hong, Xiaoping},
  booktitle={ICRA},
  pages={5049--5055},
  year={2021},
  organization={IEEE}
}

@inproceedings{CamLiFlow,
  title={Camliflow: bidirectional camera-lidar fusion for joint optical flow and scene flow estimation},
  author={Liu, Haisong and Lu, Tao and Xu, Yihui and Liu, Jia and Li, Wenjie and Chen, Lijun},
  booktitle={CVPR},
  pages={5791--5801},
  year={2022}
}

@inproceedings{Flow2,
  title={Bring Event into RGB and LiDAR: Hierarchical Visual-Motion Fusion for Scene Flow},
  author={Zhou, Hanyu and Chang, Yi and Shi, Zhiwei},
  booktitle={CVPR},
  pages={26477--26486},
  year={2024}
}

@inproceedings{Object-Tracking1,
  title={Robust multi-modality multi-object tracking},
  author={Zhang, Wenwei and Zhou, Hui and Sun, Shuyang and Wang, Zhe and Shi, Jianping and Loy, Chen Change},
  booktitle={Proceedings of the IEEE/CVF international conference on computer vision},
  pages={2365--2374},
  year={2019}
}

@article{Object-Tracking2,
  title={Object tracking based on the fusion of roadside LiDAR and camera data},
  author={Wang, Shujian and Pi, Rendong and Li, Jian and Guo, Xinming and Lu, Youfu and Li, Tao and Tian, Yuan},
  journal={IEEE Transactions on Instrumentation and Measurement},
  volume={71},
  pages={1--14},
  year={2022},
  publisher={IEEE}
}

@inproceedings{CalibNet,
	title={CalibNet: Geometrically supervised extrinsic calibration using 3D spatial transformer networks},
	author={Iyer, Ganesh and Ram, R Karnik and Murthy, J Krishna and Krishna, K Madhava},
	booktitle={IROS},
	pages={1110--1117},
	year={2018},
	organization={IEEE}
}

@article{RGGNet,
  title={RGGNet: Tolerance aware LiDAR-camera online calibration with geometric deep learning and generative model},
  author={Yuan, Kaiwen and Guo, Zhenyu and Wang, Z Jane},
  journal={RA-L},
  volume={5},
  number={4},
  pages={6956--6963},
  year={2020},
  publisher={IEEE}
}

@inproceedings{LCCNet,
	title={LCCNet: LiDAR and camera self-calibration using cost volume network},
	author={Lv, Xudong and Wang, Boya and Dou, Ziwen and Ye, Dong and Wang, Shuo},
	booktitle={CVPR},
	pages={2894--2901},
	year={2021}
}

@inproceedings{LCCRAFT,
  title={LCCRAFT: LiDAR and Camera Calibration Using Recurrent All-Pairs Field Transforms Without Precise Initial Guess},
  author={Lee, Yu-Chen and Chen, Kuan-Wen},
  booktitle={ICRA},
  pages={16669--16675},
  year={2024},
  organization={IEEE}
}

@inproceedings{Resnet,
  title={Deep residual learning for image recognition},
  author={He, Kaiming and Zhang, Xiangyu and Ren, Shaoqing and Sun, Jian},
  booktitle={CVPR},
  pages={770--778},
  year={2016}
}

@article{SiLU,
  title={Sigmoid-weighted linear units for neural network function approximation in reinforcement learning},
  author={Elfwing, Stefan and Uchibe, Eiji and Doya, Kenji},
  journal={Neural networks},
  volume={107},
  pages={3--11},
  year={2018},
  publisher={Elsevier}
}

@inproceedings{Calib-edge1,
  title={Automatic online calibration of cameras and lasers.},
  author={Levinson, Jesse and Thrun, Sebastian},
  booktitle={Robotics: science and systems},
  volume={2},
  number={7},
  year={2013},
  organization={Citeseer}
}

@inproceedings{Calib-edge2,
  title={Online camera lidar fusion and object detection on hybrid data for autonomous driving},
  author={Banerjee, Koyel and Notz, Dominik and Windelen, Johannes and Gavarraju, Sumanth and He, Mingkang},
  booktitle={2018 IEEE Intelligent Vehicles Symposium (IV)},
  pages={1632--1638},
  year={2018},
  organization={IEEE}
}

@article{Calib-pixellevel,
  title={Pixel-level extrinsic self calibration of high resolution lidar and camera in targetless environments},
  author={Yuan, Chongjian and Liu, Xiyuan and Hong, Xiaoping and Zhang, Fu},
  journal={RA-L},
  volume={6},
  number={4},
  pages={7517--7524},
  year={2021},
  publisher={IEEE}
}

@inproceedings{Calib-MI-Grayscale,
  title={Automatic targetless extrinsic calibration of a 3d lidar and camera by maximizing mutual information},
  author={Pandey, Gaurav and McBride, James and Savarese, Silvio and Eustice, Ryan},
  booktitle={Proceedings of the AAAI conference on artificial intelligence},
  volume={26},
  number={1},
  pages={2053--2059},
  year={2012}
}

@inproceedings{MOISST,
	title={Moisst: Multimodal optimization of implicit scene for spatiotemporal calibration},
	author={Herau, Quentin and Piasco, Nathan and Bennehar, Moussab and Roldao, Luis and Tsishkou, Dzmitry and Migniot, Cyrille and Vasseur, Pascal and Demonceaux, C{\'e}dric},
	booktitle={IROS},
	pages={1810--1817},
	year={2023},
	organization={IEEE}
}

@inproceedings{SOAC,
	title={Soac: Spatio-temporal overlap-aware multi-sensor calibration using neural radiance fields},
	author={Herau, Quentin and Piasco, Nathan and Bennehar, Moussab and Roldao, Luis and Tsishkou, Dzmitry and Migniot, Cyrille and Vasseur, Pascal and Demonceaux, C{\'e}dric},
	booktitle={CVPR},
	pages={15131--15140},
	year={2024}
}

@inproceedings{3DGS-Calib,
	title={3dgs-calib: 3d gaussian splatting for multimodal spatiotemporal calibration},
	author={Herau, Quentin and Bennehar, Moussab and Moreau, Arthur and Piasco, Nathan and Rold{\~a}o, Luis and Tsishkou, Dzmitry and Migniot, Cyrille and Vasseur, Pascal and Demonceaux, C{\'e}dric},
	booktitle={IROS},
	pages={8315--8321},
	year={2024},
	organization={IEEE}
}

@inproceedings{CalibDepth,
  title={Calibdepth: Unifying depth map representation for iterative lidar-camera online calibration},
  author={Zhu, Jiangtong and Xue, Jianru and Zhang, Pu},
  booktitle={ICRA},
  pages={726--733},
  year={2023},
  organization={IEEE}
}

@article{CorrI2P,
  title={CorrI2P: Deep image-to-point cloud registration via dense correspondence},
  author={Ren, Siyu and Zeng, Yiming and Hou, Junhui and Chen, Xiaodong},
  journal={IEEE Transactions on Circuits and Systems for Video Technology},
  volume={33},
  number={3},
  pages={1198--1208},
  year={2022},
  publisher={IEEE}
}

@article{CoFiI2P,
  title={CoFiI2P: Coarse-to-Fine Correspondences-Based Image to Point Cloud Registration},
  author={Kang, Shuhao and Liao, Youqi and Li, Jianping and Liang, Fuxun and Li, Yuhao and Zou, Xianghong and Li, Fangning and Chen, Xieyuanli and Dong, Zhen and Yang, Bisheng},
  journal={RA-L},
  year={2024},
  publisher={IEEE}
}

@article{NerF,
  title={Nerf: Representing scenes as neural radiance fields for view synthesis},
  author={Mildenhall, Ben and Srinivasan, Pratul P and Tancik, Matthew and Barron, Jonathan T and Ramamoorthi, Ravi and Ng, Ren},
  journal={Communications of the ACM},
  volume={65},
  number={1},
  pages={99--106},
  year={2021},
  publisher={ACM New York, NY, USA}
}

@article{RoPE,
  title={Roformer: Enhanced transformer with rotary position embedding},
  author={Su, Jianlin and Ahmed, Murtadha and Lu, Yu and Pan, Shengfeng and Bo, Wen and Liu, Yunfeng},
  journal={Neurocomputing},
  volume={568},
  pages={127063},
  year={2024},
  publisher={Elsevier}
}

@article{NaViT,
  title={Patch n’pack: Navit, a vision transformer for any aspect ratio and resolution},
  author={Dehghani, Mostafa and Mustafa, Basil and Djolonga, Josip and Heek, Jonathan and Minderer, Matthias and Caron, Mathilde and Steiner, Andreas and Puigcerver, Joan and Geirhos, Robert and Alabdulmohsin, Ibrahim M and others},
  journal={NeurIPS},
  volume={36},
  pages={2252--2274},
  year={2023}
}

@inproceedings{KITTI,
    title={Are we ready for autonomous driving? the kitti vision benchmark suite},
    author={Geiger, Andreas and Lenz, Philip and Urtasun, Raquel},
    booktitle={CVPR},
    pages={3354--3361},
    year={2012},
    organization={IEEE}
}

@article{nuScenes,
  title={nuScenes: A multimodal dataset for autonomous driving},
  author={Holger Caesar and Varun Bankiti and Alex H. Lang and Sourabh Vora and 
          Venice Erin Liong and Qiang Xu and Anush Krishnan and Yu Pan and 
          Giancarlo Baldan and Oscar Beijbom},
  journal={arXiv preprint arXiv:1903.11027},
  year={2019}
}

@article{ViT,
  title={An image is worth 16x16 words: Transformers for image recognition at scale},
  author={Dosovitskiy, Alexey},
  journal={arXiv preprint arXiv:2010.11929},
  year={2020}
}

@inproceedings{PointNet,
  title={Pointnet: Deep learning on point sets for 3d classification and segmentation},
  author={Qi, Charles R and Su, Hao and Mo, Kaichun and Guibas, Leonidas J},
  booktitle={CVPR},
  pages={652--660},
  year={2017}
}

@article{DINOv2,
  title={Dinov2: Learning robust visual features without supervision},
  author={Oquab, Maxime and Darcet, Timoth{\'e}e and Moutakanni, Th{\'e}o and Vo, Huy and Szafraniec, Marc and Khalidov, Vasil and Fernandez, Pierre and Haziza, Daniel and Massa, Francisco and El-Nouby, Alaaeldin and others},
  journal={arXiv preprint arXiv:2304.07193},
  year={2023}
}

@article{PointGPT,
  title={Pointgpt: Auto-regressively generative pre-training from point clouds},
  author={Chen, Guangyan and Wang, Meiling and Yang, Yi and Yu, Kai and Yuan, Li and Yue, Yufeng},
  journal={NeurIPS},
  volume={36},
  year={2024}
}

@article{DirectCalib,
  title={General, single-shot, target-less, and automatic lidar-camera extrinsic calibration toolbox},
  author={Koide, Kenji and Oishi, Shuji and Yokozuka, Masashi and Banno, Atsuhiko},
  journal={arXiv preprint arXiv:2302.05094},
  year={2023}
}

@article{MIAS-LCEC,
  title={Online, target-free LiDAR-camera extrinsic calibration via cross-modal mask matching},
  author={Huang, Zhiwei and Zhang, Yikang and Chen, Qijun and Fan, Rui},
  journal={IEEE Transactions on Intelligent Vehicles},
  year={2024},
  publisher={IEEE}
}

@article{CalibAnything,
  title={Calib-anything: Zero-training lidar-camera extrinsic calibration method using segment anything},
  author={Luo, Zhaotong and Yan, Guohang and Li, Yikang},
  journal={arXiv preprint arXiv:2306.02656},
  year={2023}
}

@article{MSANet,
  title={MSANet: LiDAR-camera online calibration with multi-scale fusion and attention mechanisms},
  author={Xiong, Fengguang and Zhang, Zhiqiang and Kong, Yu and Shen, Chaofan and Hu, Mingyue and Kuang, Liqun and Han, Xie},
  journal={Remote Sensing},
  volume={16},
  number={22},
  pages={4233},
  year={2024},
  publisher={MDPI}
}

@inproceedings{Calibformer,
  title={Calibformer: A transformer-based automatic lidar-camera calibration network},
  author={Xiao, Yuxuan and Li, Yao and Meng, Chengzhen and Li, Xingchen and Ji, Jianmin and Zhang, Yanyong},
  booktitle={ICRA},
  pages={16714--16720},
  year={2024},
  organization={IEEE}
}
\end{document}